\documentclass{article}
\usepackage{spconf,amsmath,graphicx,hyperref}

\usepackage{amsthm, amssymb}
\usepackage{enumitem}

\usepackage{algorithm}
\usepackage{algorithmic}
\usepackage{cancel}
\usepackage{booktabs}
\usepackage{subcaption}

\usepackage[square,numbers]{natbib}

\hypersetup{hidelinks}

\title{Correlational Training of Morphological Neural Networks}
\name{Konstantinos Fotopoulos$^{1,2}$ \qquad Petros Maragos$^{1,2,3}$}
\address{
$^1$HERON - Hellenic Robotics Center of Excellence, Athena Research Center, Marousi, Greece \\ 
$^2$Institute of Robotics, Athena Research Center, Marousi, Greece \\
$^3$School of ECE, National Technical University of Athens, Athens, Greece \\
\texttt{k.fotopoulos@athenarc.gr}, \quad \texttt{petros.maragos@athenarc.gr}
}
\begin{document}
%
\maketitle
\begin{abstract}
Neural networks are typically trained using first-order methods and back-propagation. It is unclear whether this approach is optimal for morphological layers whose weight Jacobians are sparse and whose resulting parameter gradients can be poor. In this work, we propose a novel weight update method for morphological neural networks inspired from the Multiplicative Weights Update (MWU) scheme. We view each morphological perceptron as an instance of the learning from experts' advice problem in logarithmic space, and use a correlation-based reward that favors inputs aligned with the desired output change, regardless of whether a strong gradient signal has reached their weight. We empirically evaluate our approach by training fully connected layers both as stand-alone models and as parts of larger transformer networks. Across nine benchmarks, correlational training yields improvements on eight, by up to 32.84 percentage points, while substantially reducing run-to-run variability\footnote{Code available: https://github.com/kostfoto/Morph-MWU.git}. 
\end{abstract}
\section{Introduction}\label{sec:intro}
First-order optimization methods such as stochastic gradient descent and back-propagation have enabled efficient training of typical neural networks predominantly based on matrix multiplications combined with nonlinear activations \cite{goodfellow2016deep}. It is, however, not obvious that this approach is optimal for alternative architectures with sparse or otherwise poorly informative parameter Jacobians. Among them, morphological networks 
\cite{davidson1993morphology, yang1995minmax, ritter1996introduction, pessoa2000neural} are of particular interest due to their lack of multiplications --- making them potentially more inference-efficient than their linear counterparts --- and their rich supporting theoretical foundation 
\cite{haralick1987image, heihmans1994morphological, maragos1987morphological, serra1982image}. These networks are based on the morphological operations of dilation and erosion, replacing addition by maximum and multiplication by addition. Modern architectures include the Min-max-plus network of \cite{luo2021min} and MPM networks of \cite{fotopoulos2025training}, which are based on the representation theory of mathematical morphology \cite{maragos1987morphological, banon1993decomposition} and come with theoretical guarantees. 

Morphological networks \cite[Chap.4]{Mara25} are notoriously difficult to train. As shown in \cite{dimitriadis2021advances, dimitrova2026learning, fotopoulos2025training}, this difficulty stems from the sparsity of their gradient signals. In addition, \cite{fotopoulos2025training} argues that the non-informativeness of the gradient is also a limiting factor. \cite{fotopoulos2025training} tackles this by using \emph{depth} to allow the gradient signal to disperse across all variables of the input. However, in many modern applications the feed-forward components comprise very shallow blocks of the network; see for instance the fully connected layers of transformer blocks. From the representation theorem of mathematical morphology \cite{maragos1987morphological, banon1993decomposition}, expressed in its discrete form \cite[Thm. 4.2]{luo2021min}, we know that shallow networks of suprema of erosions should be expressive, as they are zeroth-order universal approximators. Although such networks can approximate function values, their derivatives need not approximate those of the target function, inhibiting efficient training. This leaves open the question: Can we reliably train shallow Min-max-plus networks despite the sparsity and non-informativeness of their gradient signals? 

Normalization techniques and/or dropout \cite{luo2021min, zhang2019maxplus, fotopoulos2025training} have been used to mitigate the issue. Smoothing techniques \cite{masci2013learning} change the forward operator so that ordinary chain-rule gradients become denser. We also work with a softened morphological operator, but rather than relying solely on the denser chain-rule gradients produced by smoothing, we replace the morphological weight update itself using a surrogate reward. Alternative optimization methods include the use of the convex-concave procedure for training dilation-erosion perceptrons \cite{charisopoulos2017morphological} and min-max classifiers \cite{valle2026training}, but are limited to smaller networks. Our contributions are as follows: 

\begin{itemize}[left=0pt, nosep]
    \item We frame the training of the weights of a softened morphological perceptron as a learning from experts' advice problem in logarithmic space. 
    \item We propose the first, to the best of our knowledge, correlation-based, MWU-inspired update for morphological weights that bypasses their sparse local Jacobian.
    In particular, we put forward a weight update method that rewards inputs that correlate well with the desired change in the output, regardless of whether they see a strong gradient signal. 
    \item We evaluate our method empirically by training fully connected layers both as stand-alone networks on simple classification tasks, and as components of larger transformers.
\end{itemize}

\section{Prerequisites and notation}\label{sec:prereq}

Mathematical morphology \cite{heihmans1994morphological, serra1988image} is built upon complete lattices, which are defined as partially ordered sets \((\mathbb{L}, \preceq)\) for which every subset \(X \subseteq \mathbb{L}\) has a supremum \(\bigvee X\) and an infimum \(\bigwedge X\). Morphology studies operators between lattices. 
Let \(\overline{\mathbb{R}}=\mathbb{R}\cup\{+\infty, -\infty\}\) be the extended reals. The vector set \(\overline{\mathbb{R}}^n\), equipped with the partial order \(\mathbf{x} \preceq \mathbf{y} \Leftrightarrow x_i \leq y_i,\forall i\), forms a complete lattice.
(Anti-)dilations and (anti-)erosions from \(\overline{\mathbb{R}}^n\) to \(\overline{\mathbb{R}}\) can be defined as
\begin{align}
\delta_{\mathbf{w}}(\mathbf{x}) &= \max_{i=1}^n (x_i + w_i), \ \varepsilon_{\mathbf{m}}(\mathbf{x}) = \min_{i=1}^n (x_i + m_i) \label{eq:dilation-erosion}, \\
\delta^a_{\mathbf{m}}(\mathbf{x}) &= \min_{i=1}^n (-x_i + m_i), \ \varepsilon^a_{\mathbf{w}}(\mathbf{x}) = \max_{i=1}^n (-x_i + w_i). \label{eq:anti}
\end{align} 

Combining these building blocks as perceptrons, we can construct more complex functions, guaranteed by the representation theory of mathematical morphology \cite{maragos1987morphological, banon1993decomposition}: if \(f:\mathbb{R}^n \to \mathbb{R}\) is a 1-Lipschitz function, we can write it as a supremum over a combined erosion and anti-dilation:
\[
f(\mathbf{x}) = \sup_{\mathbf{r} \in \mathbb{R}^n} \left[f(\mathbf{r}) + (\varepsilon_{\mathbf{r}}(\mathbf{x}) \wedge \delta^a_{-\mathbf{r}}(\mathbf{x}))\right]. 
\]
By discretizing the representation theorem, we obtain an approximation of \(f\) in the form of a Min-max-plus network \cite{luo2021min}: 
\begin{equation*}
f(\mathbf{x})\! \approx\! \max_{i} \left[f(\mathbf{r}_i)\! +\! (\min_{j}(x_{j} + r_{i,j}) \wedge \min_{j}(-x_{j} - r_{i,j}))\right].
\end{equation*}
Training such a network using gradient descent can be troublesome due to the sparsity of the gradient signal during back-propagation and the non-first-order universality of the network \cite{dimitriadis2021advances, dimitrova2026learning, fotopoulos2025training}. The solution of \cite{fotopoulos2025training} makes use of depth, which we here aim to keep small, as is usual for fully connected components.

Finally, in the learning from experts' advice problem, we seek a distribution \(\{p_t(i)\}_{i \in I}\) minimizing the regret of mixed predictions \(y_t=\sum_i p_t(i) x_t(i)\) over given experts' opinions \(\{x_t(i)\}_{i \in I}\). The Multiplicative Weights Update (MWU) \cite{arora2012multiplicative} method solves this problem by rewarding well-performing experts via a reward \(r_t(i)\), updating expert's \(i\) weight \(p_{t+1}(i) \propto p_t(i)e^{\gamma r_t(i)}\) and re-normalizing. One of our contributions lies in the construction of the reward model \(r_t(i)\). 

\section{Method}
Consider a dilation \(\delta_{\tilde{\mathbf{w}}}(\mathbf{x}) = \bigvee_{i=1}^n (x_i+\tilde w_i)\) with inputs \(\mathbf{x} \in \mathbb R^n\) and weights \(\tilde{\mathbf{w}} \in \mathbb{R}^n\), as in \eqref{eq:dilation-erosion}. Its subdifferential with respect to the weights is given by
\[
\partial \delta_{\tilde{\mathbf{w}}}(\mathbf{x}) = \operatorname{conv}(\{\mathbf{e}_i : i \in \operatorname{argmax}_i(x_i+\tilde w_i)\}). 
\]
\(\delta_{\tilde{\mathbf{w}}}(\mathbf{x})\) is a Lipschitz continuous function, and thus differentiable almost everywhere, and precisely when the \(\operatorname{argmax}\) in the equation of the subdifferential is a singleton. In this case, all but one weights receive a gradient of \(0\), and optimization using gradient descent becomes inefficient. 

Notice that a dilation as in \eqref{eq:dilation-erosion} is essentially a competition among the inputs, with some a priori bias given to each input in the form of a weight \(w_i\). Our method is motivated by the Maslov dequantization of this maximum operator. We consider functions of the form
\begin{equation*}
y_\tau(\mathbf{x}, \mathbf{w}, c) = c + \tau \log\sum_{i=1}^n p_i e^{x_i / \tau}, \ p_i = e^{w_i / \tau},  
\end{equation*}
with weights \(\mathbf{w}\) such that \(\sum_i p_i = \sum_i e^{w_i/\tau} = 1\).
Here, \(\{x_i\}_{i=1}^n\) function as experts' opinions in log-space. \(\{p_i\}_{i=1}^n\) is a distribution over the inputs, with \(\{w_i\}_{i=1}^n\) the experts' weights in log-space. \(c\) is a common output translation that restores the common additive degree of freedom removed by normalizing the expert weights. \(\tau\) controls the softness. In particular, as \(\tau \to 0^+\), \(y\) reduces to a dilation with unconstrained weights \(\tilde{\mathbf{w}} = \mathbf{w} + c\mathbf{1}\):
\[
y_\tau(\mathbf{x}; \mathbf{w}, c) \xrightarrow{\tau \to 0} \max_{i=1}^n (x_i + w_i + c) = \max_{i=1}^n (x_i + \tilde{w}_i). 
\]

Suppose we are given the inputs over a batch \(\mathcal{B}\), and that we can obtain some desired change \(d\) in the outputs over \(\mathcal B\). This can be obtained, for instance, via back-propagation as \(d_{y(\mathbf{x}; \mathbf{w}, c)} = -\nabla_{y(\mathbf{x}; \mathbf{w}, c)} \mathcal{L}\), where \(\mathcal{L}\) is a loss that depends on the layer-output \(y(\mathbf{x}; \mathbf{w}, c)\). We decompose the desired change as follows: 
\begin{align}\label{eq:decomposition}
d_{y(\mathbf{x}; \mathbf{w}, c)} &= \underbrace{\mathbb E_{\mathcal B} [d(\mathbf{x}; \mathbf{w}, c)]}_{\text{handled by } c} + \underbrace{(d_{y(\mathbf{x}; \mathbf{w}, c)} - \mathbb E_{\mathcal B} [d(\mathbf{x}; \mathbf{w}, c)])}_{\text{handled by experts' } w_i}
\end{align}
The first term is a uniform desired drift in the values that the output takes. Since \(\partial_c y = 1\), the common translation \(c\) can realize the uniform drift exactly. Thus, we apply a preconditioning step by updating \(c\) to follow the common drift, and the weights \(w_i\) to follow the centered desired change \(\tilde d = d - \mathbb E_{\mathcal B}[d]\). Hence, we update \(c\) as
\begin{equation}\label{eq:update_scaling}
c' = c + \gamma \mathbb{E}_{\mathcal B} [d_{y(\mathbf{x}; \mathbf{w}, c)}]. 
\end{equation}
Setting \(\bar y = y - c\), differentiating the loss with respect to the expert probability \(p_i\), using the chain-rule, and preconditioning by propagating the centered \(\tilde d\) instead of \(d\) gives the following signal of desired change in \(p_i\):
\begin{equation*}
s_i = \tilde d_{y(\mathbf{x}; \mathbf{w}, c)} \tau \frac{e^{x_i/\tau}}{\sum_j p_j e^{x_j/\tau}} = \tilde d_{y(\mathbf{x}; \mathbf{w}, c)} \tau e^{(x_i - \bar y)/\tau}.
\end{equation*}
We apply a Taylor expansion to the exponential to get
\begin{align*}
s_i = \tilde d_{y(\mathbf{x}; \mathbf{w}, c)} \left(\tau + (x_i - \bar y) + O\left(\frac{(x_i - \bar y)^2}{\tau}\right)\right).
\end{align*}
Dropping the higher order term --- which becomes increasingly accurate as \(|x_i - \bar y|/\tau\) decreases --- and taking expectation yields a linear surrogate of the preconditioned signal of desired change in \(p_i\), which we use as the reward model of an exponentiated-gradient/MWU scheme:
\begin{equation}
r(i) = \mathbb{E}_{\mathcal B}[s_i] = \underbrace{\tau \mathbb{E}_{\mathcal B} [\tilde d_{y(\mathbf{x}; \mathbf{w}, c)}]}_{=0} + \mathbb{E}_{\mathcal B} [\tilde d_{y(\mathbf{x}; \mathbf{w}, c)}(x_i - \bar y)],  \label{eq:total_exp_loss}
\end{equation}
The reward measures whether expert \(i\)'s advice varies across the batch in the same direction as the desired change of the layer-output. In probability-space, the expert probability is updated according to the above reward via a canonical exponentiated-gradient/MWU update step as
\begin{equation}\label{eq:update_prob}
p_i' \propto p_i e^{\gamma r(i)} = p_i e^{\gamma \mathbb{E}_{\mathcal B} [\tilde d_{y(\mathbf{x}; \mathbf{w}, c)}(x_i - \bar y)]}, \ \sum_i p_i' = 1. 
\end{equation}
Note that \(\mathbb{E}_{\mathcal B}[-\tilde d_{y(\mathbf{x}; \mathbf{w}, c)} \bar y]\) is constant among the experts, and MWU is invariant to adding an expert-independent constant to all rewards. Hence, because \(\mathbb{E}_{\mathcal B}[\tilde d] = 0\), the method is equivalent to using the reward
\[
r'(i) = \mathbb E_{\mathcal B} [\tilde d_{y(\mathbf{x}; \mathbf{w}, c)} x_i] = \operatorname{Cov}(\tilde d_{y(\mathbf{x}; \mathbf{w}, c)}, x_i), 
\]
which justifies it being termed \emph{correlational training}. 

Returning back to log-space, \eqref{eq:update_prob} corresponds to the following update of the weight \(w_i\): 
\begin{align}\label{eq:update_weight}
& \frac{w_i'}{\tau} \leftarrow \frac{w_i}{\tau} + \gamma r(i) \notag \\ 
\Leftrightarrow \ & w_i' \leftarrow w_i + \tau \gamma r(i), \ \tau\log\sum_ie^{w_i' /\tau} = 0, 
\end{align}
where we can absorb \(\tau\) into the step-size \(\gamma\). 

For hard morphology (\(\tau=0\)), we use the corresponding zero-temperature limit: the same surrogate rewards update the morphological weights additively, followed by max-normalization \(w_i \leftarrow w_i - \max_j w_j\). 

For complex networks, we can run back-propagation as normal, retaining the inputs, outputs, and gradients of all intermediate dilations. Note that erosions, anti-dilations and anti-erosions can be expressed via dilations with negations of inputs and/or outputs. Then, we can update all non-morphological parameters of the network as normal, and the morphological parameters based on our MWU scheme. In practice, we also apply acceleration methods such as momentum or Adam over our MWU updates. We summarize our training method in Algorithm~\ref{alg:method}.

\begin{algorithm}[t]
    \centering
    \caption{MWU-adjusted morphological training.}
    \label{alg:method}
    \begin{algorithmic}[1]
        \STATE Run back-propagation to obtain the intermediate desired changes \(d_y = -\nabla_{y} \mathcal L\) over the batch \(\mathcal B\) of all dilations \(\delta_{\mathbf{w}}(\mathbf x)\) comprising the network. Also maintain their inputs \(\mathbf x\) and outputs \(y = \delta_{\mathbf{w}}(\mathbf{x})\) across the batch \(\mathcal B\). 
        \STATE For all non-morphological parameters \(\theta\) of the network, set \(g_{\theta} = -\nabla_\theta \mathcal{L}\). 
        \STATE For common output translation parameters \(c\), set \(g_c = \mathbb{E}_{\mathcal B} [d_y]\).
        \STATE For morphological weights \(w_i\) set the reward \(g_{w_i} = r(i) = \mathbb{E}_{\mathcal B} [\tilde d_{y}(x_i - \bar y)]\), where \(\tilde d_{y} = d_y - \mathbb{E}_{\mathcal B}[d_y]\) and \(\bar y = y - c\). 
        \STATE For some step-size \(\eta\), update all weights \(\theta\) of the network according to 
        \[
        \theta' = \theta + \eta g_{\theta}.
        \] 
        \vspace{-\baselineskip}
        \STATE For all morphological weights \(w_i\), re-normalize: 
        \[
        w_i \leftarrow w_i - \tau \log\sum_j\exp(w_j / \tau). 
        \]
        \vspace{-\baselineskip}
    \end{algorithmic}
\end{algorithm}

\section{Experiments}

\begin{table*}[t]
    \centering
    \caption{Test performance of linear models, morphological models trained using ordinary back-propagation (BP), and the proposed correlational update, both accelerated using Adam, across several benchmarks. The fourth column corresponds to percentage-point difference with respect to ordinary BP. For image datasets we report accuracy; for PMLB balanced accuracy.}
    \label{tab:results}
    \begin{tabular}{lcccc}
    \toprule
    Dataset & MLP / ViT & Morph. + BP & Morph. + Corr. (ours) & $\Delta$ vs. BP (pp) \\
    \midrule
    CIFAR-10 & $81.08 \pm 0.33$ & $63.96 \pm 0.88$ & $70.20 \pm 0.62$ & +6.24 \\
    Fashion-MNIST & $88.94 \pm 0.34$ & $62.66 \pm 8.18$ & $86.03 \pm 0.36$ & +23.37 \\
    MNIST & $98.19 \pm 0.20$ & $62.46 \pm 8.90$ & $95.30 \pm 0.38$ & +32.84 \\
    Adult & $78.99 \pm 1.15$ & $59.14 \pm 12.52$ & $78.13 \pm 1.08$ & +18.99 \\
    Chess & $98.96 \pm 0.33$ & $68.70 \pm 8.29$ & $87.89 \pm 12.69$ & +19.19 \\
    Ionosphere & $90.48 \pm 2.94$ & $78.38 \pm 16.03$ & $83.65 \pm 4.74$ & +5.27 \\
    Mushroom & $99.98 \pm 0.05$ & $73.01 \pm 21.35$ & $96.32 \pm 2.57$ & +23.31 \\
    Phoneme & $84.21 \pm 0.81$ & $67.58 \pm 11.61$ & $78.45 \pm 2.65$ & +10.87 \\
    Spambase & $92.98 \pm 0.81$ & $87.51 \pm 2.53$ & $87.36 \pm 1.99$ & -0.15 \\
    \bottomrule
    \end{tabular}
\end{table*}

\begin{figure}[t]
    \centering
    \begin{subfigure}[b]{0.45\linewidth}
        \centering
        \includegraphics[width=\textwidth]{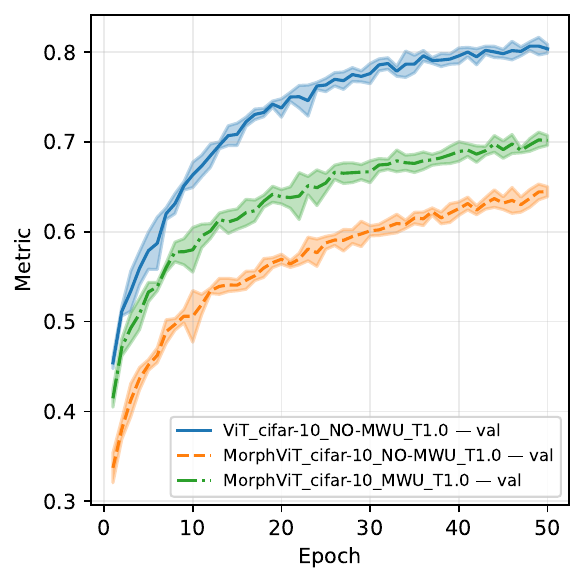}
        \caption{Transformers, CIFAR-10}
        \label{fig:training_curve_mnist}
    \end{subfigure}
    \begin{subfigure}[b]{0.45\linewidth}
        \centering
        \includegraphics[width=\textwidth]{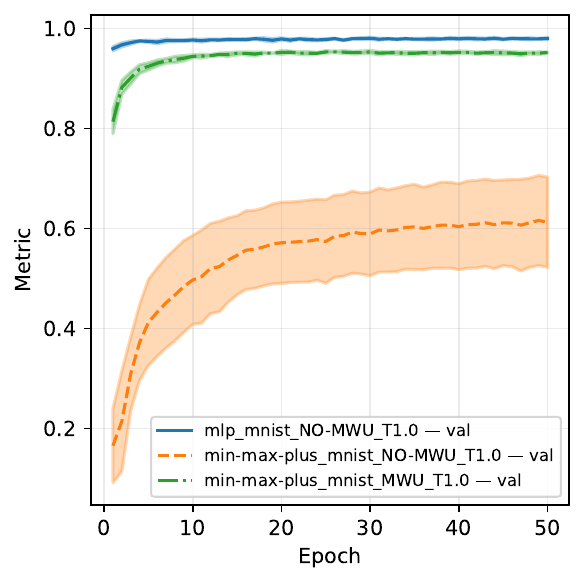}
        \caption{Fully connected, MNIST}
        \label{fig:training_curve_cifar}
    \end{subfigure}
    \caption{Training curves (mean $\pm$ std) of linear models (blue), morphological models under vanilla Adam (orange), and morphological models under our proposed method (green). Our method improves training dynamics over vanilla Adam.}
    \label{fig:training_curves}
\end{figure}

\textbf{Experimental setup.}
We evaluate our method on six PMLB classification benchmarks \cite{olson2017pmlb}, MNIST and Fashion-MNIST \cite{deng2012mnist}, and CIFAR-10 \cite{krizhevsky2009learning}. On PMLB, we use single-block hard (\(\tau=0\)) Min-max-plus networks of width \(128\); on (Fashion-)MNIST, we use two-block soft (\(\tau=1\)) networks of width \(256\). All blocks are implemented in max--min order and operate on signed inputs \((x,-x)\) with learnable coordinate multipliers following \cite{luo2021min}. On CIFAR-10, we use a six-block ViT with patch size \(4\), embedding dimension \(128\), and feed-forward width \(256\); the morphological ViT replaces its feed-forward modules by corresponding soft min-max morphological modules while leaving the remaining transformer architecture unchanged. ReLU MLPs and the standard ViT serve as parameter-matched conventional baselines, while our primary controlled comparison is between identical morphological models trained with vanilla back-propagation and with the proposed correlational update.

All models use Adam with zero weight decay and cross-entropy loss as a black-box accelerator. We train for \(100\) epochs on PMLB and \(50\) epochs on the image datasets, with batch sizes \(32\) and \(64\), respectively. Learning rates are selected for stable training: \(10^{-2}\) is used for all PMLB experiments, \(10^{-3}\) for CIFAR-10, and on (Fashion-)MNIST \(10^{-2}\) for correlational training and \(10^{-3}\) otherwise. We report accuracy on the image datasets and balanced accuracy on PMLB. Image training sets are split \(85/15\%\) into training and validation sets while retaining the official test sets; PMLB uses stratified \(65/15/20\%\) train/validation/test splits. PMLB features are standardized using training-set statistics, and CIFAR-10 uses standard random crop and horizontal-flip augmentation during training only. We select the checkpoint with the best validation metric and report test mean and standard deviation over five seeds (\(42\)-\(46\)); for PMLB, each seed determines a split shared by all compared methods.

\textbf{Main results.}
Table~1 reports test performance across all benchmarks. \emph{Correlational training improves the morphological model over vanilla back-propagation on eight of the nine tasks}, with gains ranging from \(5.27\) to \(32.84\) percentage points, while performance on Spambase remains essentially unchanged. The largest gains occur on MNIST (\(+32.84\) pp), Fashion-MNIST (\(+23.37\) pp), Mushroom (\(+23.31\) pp), Chess (\(+19.19\) pp), and Adult (\(+18.99\) pp). On Adult, the proposed method nearly closes the gap to the MLP baseline (\(78.13\%\) vs.\ \(78.99\%\)); similarly, on MNIST and Fashion-MNIST it recovers most of the performance lost when the same morphological architectures are trained with ordinary back-propagation. The proposed training rule substantially closes the gap to MLP/ViT, but does not generally eliminate it; the conventional MLP/ViT baselines remain stronger on most tasks, indicating that our contribution primarily addresses the trainability of morphological architectures, providing a promising first step torward closing this gap. 

\textbf{Training dynamics.}
Figure~1 illustrates the optimization behavior on MNIST and CIFAR-10. On MNIST, vanilla training of the Min-max-plus network converges slowly and showcases substantial dependence on the chosen seed, whereas correlational training rapidly reaches high accuracy and remains stable across runs. The same qualitative behavior persists when min-max-plus layers replace the feed-forward component of transformer blocks: on CIFAR-10, the morphological ViT improves from \(63.96\pm0.88\%\) under vanilla training to \(70.20\pm0.62\%\) under the proposed correlational update. The reduction in variance is also apparent across several PMLB datasets; for example, the standard deviation decreases from \(12.52\) to \(1.08\) percentage points on Adult and from \(21.35\) to \(2.57\) on Mushroom. These results suggest that the proposed update not only improves final performance, but also reduces the tendency of optimizing morphological layers to converge to poorly performing solutions.

\textbf{Limitations.} A remaining challenge is the hard limit \(\tau \to 0\), where morphology becomes increasingly sparse and harder to optimize. Our method is beneficial in this regime, as shown by the hard PMLB experiments, but on image classification the softened \(\tau = 1\) setting was more competitive. Thus, bridging soft and hard morphology remains an open optimization challenge rather than one specific our method.

\section{Conclusion}
We proposed a correlation-based weight update rule for morphological neural networks based on a learning-from-experts interpretation of softened morphological perceptrons. Instead of relying on the sparse local Jacobian of the morphological weights, our method uses a centered layer-output gradient and its correlation with the activations of each input to construct expert rewards, and updates normalized log-weights using an EG/MWU-inspired rule. Across stand-alone Min-max-plus networks and morphological transformer blocks, the method substantially improves both performance and training stability over ordinary back-propagation, with gains on eight of the nine tested benchmarks. While conventional MLP and ViT baselines remain stronger on most tasks, our results indicate that poor first-order trainability can substantially underestimate the practical capability of morphological architectures.

\bibliographystyle{IEEEbib_compact}
\bibliography{refs}

\end{document}